\pdfoutput=1
\RequirePackage[hyphens]{url}

\documentclass[11pt]{article}

\usepackage{acl}

\usepackage{times}
\usepackage{latexsym}
\usepackage{graphicx}
\usepackage{tabularx}
\usepackage{listings}
\usepackage{xcolor}
\usepackage{multicol}
\usepackage[linesnumbered, plainruled]{algorithm2e}
\usepackage{algpseudocode}
\usepackage[font=footnotesize]{subcaption}
\usepackage{footnote}
\makesavenoteenv{algorithm}
\usepackage{pgfplots}
\SetAlCapNameFnt{\scriptsize}
\SetAlCapFnt{\scriptsize}
\SetAlgoCaptionSeparator{}
\usepackage[nameinlink]{cleveref}
\usepackage{arydshln}
\usepackage{xurl}

\makeatletter
\newcommand{\removelatexerror}{\let\@latex@error\@gobble}
\makeatother

\newenvironment{description*}%
  {\begin{description}%
    \setlength{\itemsep}{1pt}%
    \setlength{\parskip}{0pt}}%
  {\end{description}}

\makeatletter
\def\@algocf@capt@plainruled{above}
\renewcommand{\algocf@caption@plainruled}{%
  \vskip\AlCapSkip%
  \box\algocf@capbox%
  \vskip 5\algoheightrule}%
\makeatother

\definecolor{codegreen}{rgb}{0,0.6,0}
\definecolor{codegray}{rgb}{0.5,0.5,0.5}
\definecolor{codepurple}{rgb}{0.58,0,0.82}
\definecolor{backcolour}{rgb}{0.95,0.95,0.92}

\lstdefinestyle{mystyle}{
    backgroundcolor=\color{backcolour},   
    commentstyle=\color{codegreen},
    keywordstyle=\color{magenta},
    stringstyle=\color{codepurple},
    basicstyle=\ttfamily\footnotesize,
    breakatwhitespace=false,         
    breaklines=true,                 
    captionpos=b,                    
    keepspaces=true,                  
    showspaces=false,                
    showstringspaces=false,
    showtabs=false,                  
    tabsize=2
}

\usepackage[T1]{fontenc}
\usepackage[utf8]{inputenc}
\usepackage{textcomp}

\usepackage{microtype}

\title{Improving Tokenisation by Alternative Treatment of Spaces}

\author{ 
Edward Gow-Smith\textsuperscript{\textnormal{1}}, Harish Tayyar Madabushi\textsuperscript{\textnormal{2}}, \\
\textbf{Carolina Scarton\textsuperscript{\textnormal{1}}} \and
\textbf{Aline Villavicencio\textsuperscript{\textnormal{1}}}
\\[0.3cm]
\textsuperscript{1}Department of Computer Science, University of Sheffield \\
\textsuperscript{2}Department of Computer Science, University of Bath \\
\texttt{\small egow-smith1@sheffield.ac.uk}
\\
}
\begin{document}
\maketitle
\begin{abstract}
Tokenisation is the first step in almost all NLP tasks, and state-of-the-art transformer-based language models all use subword tokenisation algorithms to process input text. Existing algorithms have problems, often producing tokenisations of limited linguistic validity and representing equivalent strings differently depending on their position within a word. We hypothesise that these problems hinder the ability of transformer-based models to handle complex words, and suggest that these problems are a result of allowing tokens to include spaces. We thus experiment with an alternative tokenisation approach where spaces are always treated as individual tokens. Specifically, we apply this modification to the BPE and Unigram algorithms. We find that our modified algorithms lead to improved performance on downstream NLP tasks that involve handling complex words, whilst having no detrimental effect on performance in general natural language understanding tasks. Intrinsically, we find that our modified algorithms give more morphologically correct tokenisations, in particular when handling prefixes. Given the results of our experiments, we advocate for always treating spaces as individual tokens as an improved tokenisation method.
\end{abstract}

\section{Introduction}
Tokenisation is a key initial step in processing natural language, as it identifies the linguistic units to be processed, converting them to numerical IDs which can then be vectorised and manipulated by mathematical operations.

Earlier NLP approaches used simple string-searching techniques with regular expressions to tokenise text; however, these pattern-matching tokenisation methods have drawbacks: they require large vocabulary sizes to cover the training data, they cannot handle out-of-vocabulary words, and they do not work for languages without spaces as word boundaries. To address these issues, subword tokenisation was introduced. The first explicit mention (and popularisation) of this approach was by ~\citet{sennrich2015neural}, though it was indirectly introduced earlier by ~\citet{schuster2012japanese}. This method works by learning from training data to build a vocabulary (of a fixed size) and then tokenising text at inference time using this vocabulary (and possibly other learnt parameters). More frequent words are represented as single tokens, with rare words being broken down into multiple subword tokens, possibly down to the character level. 

State-of-the art transformer-based language models all use subword tokenisation algorithms based on either byte-pair encoding (BPE) \cite{sennrich2015neural} or Unigram \cite{kudo2018subword}.
The original transformer model ~\cite{vaswani2017attention} uses BPE, whilst BERT ~\cite{devlin2018bert}, which consists of a transformer encoder pretrained with a masked language modelling objective, uses WordPiece tokenisation ~\cite{schuster2012japanese}, which is a variant of BPE with a language model loss function. WordPiece is also used by ERNIE ~\cite{sun2019ernie}, DistilBERT ~\cite{sanh2019distilbert}, ELECTRA ~\cite{clark2020electra}, StructBERT ~\cite{wang2019structbert} and NEZHA ~\cite{wei2019nezha}.  
GPT-2 ~\cite{radford2019language} introduced byte-level BPE, operating on byte sequences rather than Unicode code points, which allows all sequences to be encoded using a base vocabulary of 256, avoiding the issue of unknown characters. The same approach is used in RoBERTa ~\cite{liu2019roberta}, DeBERTa ~\cite{he2020deberta}, and BART ~\cite{lewis-etal-2020-bart}. 

The BPE and Unigram algorithms are implemented in the SentencePiece library ~\cite{kudo2018sentencepiece}. There is a lack of clarity regarding SentencePiece in the literature, with it being erroneously considered as its own algorithm rather than an implementation of other algorithms. For example, in the paper introducing T5 ~\cite{raffel2019exploring} they state that they "use SentencePiece to encode text as WordPiece tokens", which is not in fact implemented in SentencePiece. Looking at their code, we find that they use the default SentencePiece implementation, which is Unigram. XLNET ~\cite{yang2019xlnet} say they tokenise with SentencePiece, but do not say which algorithm they use - again, looking at their code, we find they use the default of Unigram. Equivalently, ALBERT \cite{lan2019albert} say that they tokenise with SentencePiece as for XLNET, meaning they again use Unigram.  

Despite their ubiquity, existing tokenisation algorithms have problems, which we hypothesise hinders the ability of language models to handle complex words (\Cref{sec:problems-with-existing-algorithms}). We suggest that these problems are pervasive across all existing subword tokenisation algorithms due to a shared fundamental design choice of allowing tokens to include spaces, and thus experiment with an alternative treatment of spaces where they are always taken as individual tokens. We implement this approach by making simple modifications to the existing WordPiece, BPE, and Unigram algorithms (\Cref{sec:modified-algorithms}). We first evaluate our modified algorithms intrinsically (\Cref{sec:intrinsic-evaluation}), quantitatively finding that they improve morphological correctness, in particular when handling prefixes. Qualitatively, we take examples from previous papers critiquing existing tokenisation algorithms, and show how our modified algorithms are able to alleviate the discussed issues. We then evaluate our modified algorithms extrinsically by pretraining and finetuning transformer-based models (\Cref{sec:extrinsic-evaluation}), showing that they give improved performance on NLP tasks that require handling complex words with no detrimental effect on performance in the general domain.

\section{Problems with Existing Tokenisation Algorithms}
\label{sec:problems-with-existing-algorithms}

Existing tokenisation algorithms often produce unintuitive tokenisations for complex words, incorrectly splitting prefixes, and producing unmeaningful subword tokens, which are problems that have been discussed in previous works. \citet{church2020emerging} looks at the BERT (WordPiece) tokenisations for complex words, highlighting the many unnatural tokenisations that arise, with tokens often splitting up morphemes and digraphs. \citet{nayak2020domain} also discuss the issues with BERT's tokeniser, specifically highlighting problems with the splitting of prefixes, and they show that poor tokenisation leads to weak semantic representations. \citet{hofmann2021superbizarre} find that BERT performs poorly on classifying complex words containing prefixes, performing much better on suffixes. They suggest that a reason is that BERT's tokeniser is seldom accurate for splitting prefixes, but is much more often correct for splitting suffixes. \citet{schick2020rare} argue that a reason BERT struggles to understand rare words is due to suboptimal tokenisation of these words.
Here we give a few of our own examples of BERT tokenisations that illustrate the problems\footnote{BERT's tokeniser actually prepends the space symbol to subword units not occurring at the start of words, and the space symbol they use is ``\#\#'' rather than ``\_'', but these are inconsequential differences and we standardise the output here for clarity.}: 

\begin{description*}
    \small
    \item joint $\rightarrow$ \_joint
    \item jointed $\rightarrow$ \_joint, ed 
    \item disjointed $\rightarrow$ \_di, s, jo, int, ed 
    \item unisex $\rightarrow$ \_un, ise, x 
    \item true $\rightarrow$ \_true 
    \item untrue $\rightarrow$ \_un, tr, ue 
    \item estimate $\rightarrow$ \_estimate 
    \item overestimate $\rightarrow$ \_over, est, imate 
\end{description*}

We see here that the prefixed words are tokenised poorly: the prefix is either incorrectly split, as in ``disjointed'' and ``unisex'', or the prefix is correctly split, but the rest of the word is tokenised differently from the standalone case, as in ``untrue'' and ``overestimate''. We note that suffixes are handled better than prefixes, which is due to spaces being prepended rather than appended to words (see \Cref{sec:modified-algorithms}).

For these latter examples, there is a second problem: even if the base were tokenised as a single token, the addition of the space symbol means that there would be no explicit link between the prefixed word and the standalone base. As an example, we cherry-pick a rare example of a morphologically correct tokenisation by BERT of a word containing a prefix, showing both strings and token IDs:

\begin{description*}
\small
\item beatable $\rightarrow$ \_beat, able (3786, 3085) 
\item unbeatable $\rightarrow$ \_un, beat, able (4895, 19442, 3085)
\end{description*}

We can see that, even though these tokenisations are reasonable, the subword ``beat'' is assigned different IDs in the two cases due to the prepending of the special space symbol. 

We hypothesise that both of these problems hinder the ability of existing language models (such as BERT) to deal with complex words. Regarding the first problem, we argue that the morphological correctness of a tokeniser is a metric which will correlate with the ability of language models to deal with complex words: correctly splitting affixes means morphologically related words (those sharing a common base) are given related tokenisations. The splitting of prefixes is particularly important, as prefixes always have a semantic function, unlike suffixes which can have both syntactic and semantic functions \cite{giraudo:hal-02434854}. Also, tokenisations made up of meaningful subword tokens (morphemes or groups of morphemes) will allow language models to build stronger representations with less data, since the representations of complex words can be computed from the representations of the subwords. Regarding the second problem, the fact that base forms are represented differently depending on their position within a word means a reduction in relevant training instances and hence a further weakening of representations for complex words.

\section{Our Modified Algorithms}
\label{sec:modified-algorithms}

We suggest that the problems discussed in \Cref{sec:problems-with-existing-algorithms} arise as a result of how spaces are handled by existing algorithms: All subword tokenisation algorithms currently used by transformer-based models allow tokens to include space symbols as the first character\footnote{Splitting on spaces occurs as a first step, so space symbols cannot occur in the middle of tokens. The default implementation splits before spaces, meaning space symbols occur only at the start of words.}. This means equivalent strings are treated differently depending on whether they appear at the start of a word or not. This difference occurs when training these tokenisers, which leads to suboptimal tokenisations of prefixed words. It also occurs when using these tokenisers in NLP models, leading to equivalent strings being assigned different tokens depending on whether they occur at the start of a word or not. 

Thus, to attempt to alleviate these issues, and hence improve the handling of complex words by language models, we propose an alternative treatment of spaces where they are always assigned individual tokens. This simple modification can be made to any existing subword tokenisation algorithm, though for brevity we focus our attention on BPE and Unigram; this modification can also be made to the WordPiece algorithm, and we see similar (intrinsic) performance improvements from doing so. In \Cref{sec:intrinsic-evaluation}, we perform a qualitative analysis of our modified WordPiece algorithm and also include the default WordPiece algorithm in our quantitative evaluation for comparison. Our modified algorithms and the defaults are shown in \autoref{fig:default-modified-bpe} and \autoref{fig:default-modified-unigram} for BPE and Unigram, respectively\footnote{We release code for training our modified algorithms, as well as running both our intrinsic and extrinsic experiments at \url{https://github.com/edwardgowsmith/improved-tokenisation-methods}}.

\begin{figure*}[ht!]
\removelatexerror
\scriptsize	
\begin{subfigure}{\textwidth}
    \begin{subfigure}[t]{.5\textwidth}
        \caption{Default BPE}\label{alg:cap}
        \begin{algorithm}[H]
        \SetKwInOut{Input}{input}
        \SetKwInOut{Output}{output}
        \Input{training data $T$, vocabulary size $s$}
        \Output{vocabulary $V$}
        Replace whitespace in $T$ with the space symbol \\
        \textcolor{red}{\textbf{Prepend the space symbol to the first word of every sentence in $T$}}\footnotemark \\
        Vocabulary $V$ initialised as all characters \\
        \While{$ |V| < s$}{
            Find the most frequently occurring bigram in $T$\textcolor{red}{\textbf{, only allowing spaces as the first character}} \\
            Apply merge operation on the bigram to make a new token \\
            Add merge operation to $V$
            }
        \caption*{Training}
        \end{algorithm}
    \end{subfigure}
    \begin{subfigure}[t]{.5\textwidth}
        \caption{Modified BPE (BPE$'$)}\label{alg:cap}
        \begin{algorithm}[H]
        \SetKwInOut{Input}{input}
        \SetKwInOut{Output}{output}
        \Input{training data $T$, vocabulary size $s$}
        \Output{vocabulary $V$}
        Replace whitespace in $T$ with the space symbol \\
        Vocabulary $V$ initialised as all characters \\
        \While{$ |V| < s$}{
            Find most frequently occurring bigram in $T$ \textcolor{green}{\emph{that does not include spaces}} \\
            Apply merge operation on the bigram to make a new token \\
            Add merge operation to $V$
            }

        \caption*{Training}
        \end{algorithm}
    \end{subfigure}
\end{subfigure}
\begin{subfigure}[b]{\textwidth}
    \begin{subfigure}[t]{.5\textwidth}
        \begin{algorithm}[H]
        \SetKwInOut{Input}{input}
        \SetKwInOut{Output}{output}
        \Input{text $T$, vocabulary $V$ }
        \Output{tokens $\tau$}
        Replace whitespace in $T$ with the space symbol \\
        \textcolor{red}{\textbf{Prepend the space symbol to the first word of every sentence in $T$}}\\ 
        Apply the merge operations from $V$ in order to $T$.
        \caption{Tokenisation}
        \end{algorithm}
    \end{subfigure}
    \begin{subfigure}[t]{.5\textwidth}
        \begin{algorithm}[H]
        \SetKwInOut{Input}{input}
        \SetKwInOut{Output}{output}
        \Input{text $T$, vocabulary $V$ }
        \Output{tokens $\tau$}
        Replace whitespace in $T$ with the space symbol \\
        Apply the merge operations from $V$ in order to $T$.
        \caption*{Tokenisation}
        \end{algorithm}
    \end{subfigure}
\end{subfigure}
\caption{Default and modified BPE algorithms. Red, bold text is removed from the default algorithm, whilst green, italic text is added. }
\label{fig:default-modified-bpe}
\end{figure*}

\begin{figure*}[ht!]
\removelatexerror
\scriptsize
\begin{subfigure}{\textwidth}
    \begin{subfigure}[t]{.5\textwidth}
        \caption{Default Unigram}\label{alg:cap}
        \begin{algorithm}[H]
        \SetKwInOut{Input}{input}
        \SetKwInOut{Output}{output}
        \Input{training data $T$, vocabulary size $s$}
        \Output{vocabulary $V$, language model parameters $\Theta$}
        Replace whitespace in $T$ with the space symbol \\
        \textcolor{red}{\textbf{Prepend the space symbol to the first word of every sentence in $T$}}\\ 
        Vocabulary $V$ initialised as all substrings occurring in $T$\textcolor{red}{\textbf{, only allowing spaces as the first character}}\footnotemark \\
        \While{$ |V| > s$}{
        Optimise a Unigram language model with parameters $\Theta$ to fit the data using the EM algorithm \\
        For each substring in $V$, compute the loss from removing this from the vocabulary \\
        Remove the substring with the smallest loss from $V$
        }
        \caption*{Training}
        \end{algorithm}
    \end{subfigure}
    \begin{subfigure}[t]{.5\textwidth}
        \caption{Modified Unigram (Unigram$'$)}\label{alg:cap}
        \begin{algorithm}[H]
        \SetKwInOut{Input}{input}
        \SetKwInOut{Output}{output}
        \Input{training data $T$, vocabulary size $s$}
        \Output{vocabulary $V$, language model parameters $\Theta$}
        Replace whitespace in $T$ with the space symbol \\
        Vocabulary $V$ initialised as all substrings occurring in $T$ \textcolor{green}{\emph{that do not include spaces, plus the space symbol}}\\
        \While{$ |V| > s$}{
        Optimise a Unigram language model with parameters $\Theta$ to fit the data using the EM algorithm \\
        For each substring in $V$, compute the loss from removing this from the vocabulary \\
        Remove the substring with the smallest loss from $V$}
        \caption{Training}
        \end{algorithm}
    \end{subfigure}
\end{subfigure}
\begin{subfigure}[b]{\textwidth}
    \begin{subfigure}[t]{.5\textwidth}
        \begin{algorithm}[H]
        \SetKwInOut{Input}{input}
        \SetKwInOut{Output}{output}
        \Input{text $T$, vocabulary $V$, language model parameters $\Theta$}
        \Output{tokens $\tau$}
        Replace whitespace in $T$ with the space symbol \\
        \textcolor{red}{\textbf{Prepend the space symbol to the first word of every sentence in $T$}}\\ 
        Use the Viterbi algorithm with the learned language modelling parameters and the vocabulary to tokenise $T$ 
        \caption{Tokenisation}
        \end{algorithm}
    \end{subfigure}
    \begin{subfigure}[t]{.5\textwidth}
        \begin{algorithm}[H]
        \SetKwInOut{Input}{input}
        \SetKwInOut{Output}{output}
        \Input{text $T$, vocabulary $V$, language model parameters $\Theta$}
        \Output{tokens $\tau$}
        Replace whitespace in $T$ with the space symbol \\
        Use the Viterbi algorithm with the learned language modelling parameters and the vocabulary to tokenise $T$ \textcolor{green}{\emph{with spaces being given an arbitrarily high score so they are always selected as individual tokens}}
        \caption{Tokenisation}
        \end{algorithm}
    \end{subfigure}
\end{subfigure}
\caption{Default and modified Unigram algorithms. Red, bold text is removed from the default algorithm, whilst green, italic text is added. }
\label{fig:default-modified-unigram}
\end{figure*}

In the following sections, we compare our modified tokenisation algorithms to the defaults by evaluating them intrinsically (\Cref{sec:intrinsic-evaluation}) and extrinsically (\Cref{sec:extrinsic-evaluation}). 

\section{Intrinsic Evaluation: Morphological Correctness}
\label{sec:intrinsic-evaluation}

Given our hypothesis that the morphological correctness of a tokeniser, especially when handling prefixes, correlates with the performance of language models in dealing with complex words (\Cref{sec:problems-with-existing-algorithms}), we perform a controlled intrinsic evaluation of our tokenisers using this metric. We train our modified algorithms and the defaults on 1 million sentences from English Wikipedia for BPE and Unigram, with a fixed vocabulary size of 16,000, and then run evaluation on four morphological datasets: LADEC, MorphoLex, MorphyNet and DagoBERT. 

The LADEC dataset \cite{gagne2019ladec} consists of 7,804 noun compounds with a unique morphological parse (we exclude those with multiple parses). MorphoLex ~\cite{sanchez2018morpholex} provides derivational morphology for 68,624 entries from the English Lexicon Project \cite{balota2007english}. Here we only consider those with a concatenative parse (i.e. no overlapping tokens), resulting in 12,028 entries. MorphyNet ~\cite{batsuren2021morphynet} provides derivational and inflectional morphology for words across 15 languages, expanding the UniMorph dataset ~\cite{mccarthy2020unimorph}. Taking only those derivational morphology entries in English with a concatenative parse gives 193,945 entries. The DagoBERT dataset ~\cite{hofmann2020dagobert} comprises 279,443 words containing low-frequency derivatives, taken from Reddit posts. Again, we take those with a concatenative parse, giving 268,513 entries. 

We evaluate a tokeniser on these datasets using the evaluation method introduced by ~\citet{creutz2004morpheme}, which produces metrics by comparing the boundaries of a generated tokenisation with a gold standard reference: false negatives are boundaries appearing in the reference but not in the generated tokenisation, whilst false positives are boundaries appearing in the generated tokenisation but not in the reference.\footnotetext[4]{In the standard implementation, space symbols are added at the start of sentences so that words are equivalent whether they appear at the start of a sentence or not.}\footnotetext[5]{This is only tractable for languages that include spaces. For languages without them, other initialisation methods must be used.} Because it makes sense to store common words as single tokens in the vocabulary, even if they can be decomposed into morphemes, we report precision along with F1 as a potentially more meaningful metric, since this allows undersegmentation whilst penalising oversegmentation. We also compute the mean sequence length (number of tokens) for each tokeniser across each dataset. Results are shown in \autoref{table:morphology-comparison}. Here, and throughout, the prime symbol ($'$) denotes the given algorithm modified to always treat spaces as individual tokens.

The general trend is that Unigram outperforms BPE (consistent with findings by \citealt{bostrom2020byte}, \citealt{hofmann-etal-2022-embarrassingly}), with the modified algorithms performing better than their default counterparts --- the average F1 scores across the four datasets are 43.0, 50.9, 59.7, and 62.4 for the four algorithms BPE, BPE$'$, Unigram, and Unigram$'$, respectively. On the MorphoLex dataset, however, the default Unigram algorithm performs the best. This is also the only dataset where default Unigram gives a shorter mean sequence length than Unigram$'$. To further investigate this, we evaluate on the subsets of the data containing only prefixed and only suffixed entries, shown in \autoref{table:morpholex-eval}. We can see that Unigram$'$ performs best on prefixed entries, but worse than default Unigram on suffixed entries. Since the dataset consists of many more entries containing suffixes than those containing prefixes (7,422 vs 2,692), this could explain the performance difference. Because the correct tokenisation of prefixed words is particularly important (\Cref{sec:problems-with-existing-algorithms}), we believe that this performance trade-off is beneficial. In \Cref{sec:extrinsic-evaluation}, we confirm this through evaluation on downstream tasks.

Interestingly, BPE$'$ gives the shortest sequence length on three of the four datasets, but not the most morphologically correct tokenisations. Since BPE was developed as a compression algorithm, the short sequence lengths are perhaps expected, but here we only see a weak correlation between sequence length and morphological correctness\footnote{Pearson's correlation of between -0.262 and -0.857 depending on the dataset.}.

\begin{table*}[hbt!]
\centering
\resizebox{\linewidth}{!}{%
\begin{tabular}{|c|c c c|c c c| c c c | c c c|}
\hline
 &  \multicolumn{3}{c|}{LADEC} &  \multicolumn{3}{c|}{MorphoLex}  &  \multicolumn{3}{c|}{MorphyNet} &  \multicolumn{3}{c|}{DagoBERT}  \\ 
 & Seq. Length & Precision  & F1 & Seq. Length & Precision & F1 & Seq. Length & Precision & F1 & Seq. Length & Precision & F1\\
\hline
WordPiece & 2.97 & 44.5 & 59.1 & 2.64 & 45.1 & 51.0 & 3.15 & 21.8 & 31.6 & 3.20 & 32.5 & 44.1 \\
\hdashline
BPE & 2.98 & 41.2 & 54.8 & 2.67 & 43.4 & 49.5 & 3.17 & 19.9 & 29.0 & 3.22 & 28.4 & 38.6 \\
BPE$'$ & 2.60 & 53.8 & 66.2 & 2.47 & 50.8 & 54.7 & 2.93 & 24.6 & 34.8 & 2.86 & 37.4 & 48.0 \\
Unigram& 2.80 & 51.9 & 66.8 & 2.56 & \textbf{58.1} & \textbf{64.3} & 3.09 & 32.3 & 46.6 & 3.16 & 45.3 & 61.1 \\
Unigram$'$ & 2.67 & \textbf{56.7} & \textbf{70.9} & 2.65 & 53.9 & 61.2 & 3.03 & \textbf{33.6} & \textbf{48.1} & 2.81 & \textbf{54.5} & \textbf{69.2} \\
\hline
\end{tabular}
}
\caption{\label{table:morphology-comparison} Performance of the tokenisation algorithms across four morphological datasets, showing the average sequence length, precision and F1 score generated following the standard introduced by \citet{creutz2004morpheme}. Best results are shown in bold.}
\end{table*}

\begin{table}[hbt!]
\centering
\resizebox{\linewidth}{!}{%
\begin{tabular}{|c|c c c|c c c|}

\hline
 & \multicolumn{3}{c|}{Only Prefixes} &  \multicolumn{3}{c|}{Only Suffixes}   \\
 &Seq. Length & Precision  & F1 & Seq. Length & Precision & F1 \\
\hline
WordPiece & 2.56 & 26.7 & 32.2 & 2.56 & 48.4 & 53.4 \\
\hdashline
BPE & 2.54 & 33.5 & 40.2 & 2.62 & 44.0 & 49.4 \\
BPE$'$ & 2.26 & 50.4 & 55.4 & 2.50 & 45.9 & 49.8 \\ 
Unigram & 2.51 & 53.4 & 63.6 & 2.53 & \textbf{55.2} & \textbf{60.5} \\ 
Unigram$'$ & 2.48 & \textbf{57.2} & \textbf{67.4} & 2.75 & 46.4 & 53.9 \\
\hline
\end{tabular}
}
\caption{\label{table:morpholex-eval} Performance of the tokenisation algorithms on subsets of the MorphoLex dataset with entries containing only prefixes and only suffixes. Best results are shown in bold.}
\end{table}

\begin{table*}[hbt!]
\resizebox{\linewidth}{!}{%
\begin{tabular}{|c|c|c|c|c|}
\hline
Input & BPE & BPE$'$ & Unigram & Unigram$'$  \\
\hline
directional & \_direction, al & direction, al & \_direction, al & direction, al\\
unidirectional & \_un, id, ire, ction, al & un, id, ire, ction, al & \_un, i, direct, ional & uni, direction, al\\ 
electroneutral & \_elect, r, one, ut, ral & electr, one, utr, al & \_electron, eu, tral & electro, neutral \\
neurotransmitter & \_neuro, trans, mit, ter & neuro, transmitter & \_neuro, trans, mitt, er & neuro, transmitter \\
responsiveness & \_respons, iveness & respons, iveness & \_re, s, pon, s, ive, ness & r, e, sp, on, s, ive, ness \\
hyporesponsiveness & \_hyp, ores, p, ons, iveness & hypo, respons, iveness & \_hypo, res, pon, s, ive, ness & hypo, r, e, sp, on, s, ive, ness \\
hyperresponsiveness & \_hyper, resp, ons, iveness & hyper, respons, iveness & \_hyper, res, pon, s, ive, ness & hyper, r, e, sp, on, s, ive, ness \\
saturated & \_sat, urated & sat, urated & \_sat, ur, ated & saturated \\
unsaturated & \_uns, atur, ated & un, sat, urated & \_un, sa, tur, ated & un, saturated \\
equal & \_equal & equal & \_equal & equal \\
unequal &
\_un, equ, al & une, qual & \_un, e, qual & un, equal \\
multiplayer & \_multip, layer & multi, player & \_multi, play, er & multi, player \\
nonmultiplayer & \_non, m, ult, ip, layer & non, multi, player & \_non, mul, ti, play, er & non, multi, player \\
overpriced & \_over, p, ric, ed & over, pr, iced & \_over, p, ric, ed & over, price, d \\
accessible & \_accessible & accessible & \_accessible & accessible \\
unaccessible & \_un, ac, cess, ible & un, accessible & \_un, ac, ces, s, ible & un, accessible \\
unicycle & \_un, icy, cle & un, icy, cle & \_un, i, cycle & uni, cycle \\
\hline
\end{tabular}}
\caption{\label{table:tokeniser-examples} Example tokenisations of the default and modified BPE and Unigram algorithms, with inputs taken from the following papers: \citet{church2020emerging}, \citet{nayak2020domain}, \citet{hofmann2020dagobert} and \citet{schick2020rare}.}
\end{table*}

\begin{table}[hbt!]
\resizebox{\linewidth}{!}{%
\begin{tabular}{|c|c|c|}
\hline
Input & WordPiece & WordPiece$'$ \\
\hline
directional & \_direction, al & direction, al \\
unidirectional & \_un, idi, rect, ional & uni, direction, al \\ 
electroneutral & \_electron, e, ut, ral & electron, eu, tra, l \\
neurotransmitter & \_ne, uro, tr, ans, mit, ter & neuro, transmit, ter \\
responsiveness & \_respons, iveness & respons, iveness \\
hyporesponsiveness & \_hyp, ores, po, n, s, iveness & hypo, respons, iveness \\
hyperresponsiveness & \_hyp, er, resp, ons, iveness & hyper, respons, iveness \\
saturated & \_sat, ura, ted & sat, urated \\
unsaturated & \_uns, at, ura, ted & un, sat, urated \\
equal & \_equal & equal \\
unequal & \_un, equ, al & une, qual \\
multiplayer & \_multip, la, yer & multi, player \\
nonmultiplayer & \_non, m, ult, ipl, ayer & non, multi, player \\
overpriced & \_over, pr, iced & over, price, d \\
accessible & \_access, ible & accessible \\
unaccessible & \_un, acc, ess, ible & una, c, cess, ible \\
unicycle & \_un, icy, cle & uni, cycle \\
\hline
\end{tabular}}
\caption{\label{table:tokeniser-examples-wordpiece} Example tokenisations of the default and modified WordPiece algorithms, with inputs taken from the following papers: \citet{church2020emerging}, \citet{nayak2020domain}, \citet{hofmann2020dagobert} and \citet{schick2020rare}.}
\end{table}

For a qualitative analysis, we take examples from papers that highlight problems with existing tokenisers (\Cref{sec:problems-with-existing-algorithms}) and generate the output from the default and modified algorithms for BPE and Unigram, shown in \autoref{table:tokeniser-examples}. These examples illustrate how our modified algorithms are able to generate improved tokenisations for complex words. For example, whereas the default Unigram algorithm tokenises ``unicycle'' into ``\_un'' ``i'' ``cycle'', which is misleading as the string ``un'' does not have its typical semantic role, our modified Unigram algorithm tokenises it more meaningfully into ``uni'' ``cycle''. Also, the modified algorithms explicitly create links between words containing prefixes and their bases. For the words ``accessible'' and ``unaccessible'', the modified algorithms tokenise the subword ``accessible'' identically in both cases. The default Unigram and BPE algorithms do correctly split the prefix ``un'', but the rest of the word is tokenised differently, which is problematic, and even if the tokenisation was equivalent, the inclusion of the space symbol means there would be no link between these forms (\Cref{sec:problems-with-existing-algorithms}). We note that our modified algorithms are not immune to oversegmentation, with Unigram$'$ tokenising ``responsiveness'' into seven tokens, although this is arguably inevitable with a limited vocabulary size. In \autoref{table:tokeniser-examples-wordpiece}, we show the same qualitative analysis between the default and modified WordPiece algorithms, finding parallels with default and modified BPE. 

We investigate the vocabularies of the default and modified algorithms, shown in \autoref{table:model-vocabs}. We remove the tokens ``[CLS]'', ``[SEP]'', and ``[UNK]'' from the vocabularies. For the default algorithms, we also remove tokens that are duplicates apart from prepended space symbols, and we find that there is significant vocabulary degeneracy (8.7\% and 9.1\% for BPE and Unigram, respectively). We also find that a large percentage of the vocabulary is transferred over from the default to the modified algorithm (90.0\% and 90.1\% for BPE and Unigram, respectively). Additionally, we see that all of the algorithms have a similar number of prefixes in their vocabularies, which suggests the tokenisation algorithm plays an important role, as performance differences on handling prefixes are large (\autoref{table:morpholex-eval}) despite similar vocabularies. This is supported by work by \citet{hofmann2021superbizarre}, who find that employing a fixed vocabulary in a morphologically correct way leads to performance improvements.  We also see, however, that Unigram$'$ has fewer suffixes in its vocabulary than default Unigram, which reflects the performance difference seen in \autoref{table:morpholex-eval}.

\begin{table}[hbt!]
\centering
\resizebox{\linewidth}{!}{%
\begin{tabular}{|c|c|c|c|c|}
\hline
 & Vocab Size & Unique Elements & \#Prefixes & \#Suffixes  \\
\hline
WordPiece & 15123 & - & 107 & 184\\
\hdashline
BPE & 14613 & 1459 & 114 & 182 \\
BPE$'$ & 15997 & 2843 & 123 & 192 \\ 
\hline
Unigram & 14544 & 1443 & 123 & 201\\ 
Unigram$'$ & 15997 & 2896 & 116 & 147 \\
\hline
\end{tabular}
}
\caption{\label{table:model-vocabs} Vocabularies of the models, showing size, number of unique elements, and numbers of prefixes and suffixes. }
\end{table}

We note that an interesting result of our modifications is an improvement at word segmentation. As an example, the outputs of the default and modified Unigram algorithms when passed the concatenated sentence ``thisisasentencethatneedstobesegmented'' are:

\begin{description}
    \small
    \item[ Unigram] \_this, isa, s, ent, ence, that, ne, ed, s, to, be, s, eg, ment, ed
    \item[ Unigram$'$] this, is, a, sentence, that, needs, to, be, segment, e, d
\end{description}

\section{Extrinsic Evaluation: Pretrain-Finetune}
\label{sec:extrinsic-evaluation}

Given the improved intrinsic performance of our algorithms, we wish to evaluate how this impacts the extrinsic performance of NLP models, both in general, and in particular on tasks involving complex words.
As in \Cref{sec:intrinsic-evaluation}, we train the default and modified BPE and Unigram algorithms on 1 million sentences from English Wikipedia, with a fixed vocabulary size of 16,000, but we also implement a variant of our modified algorithm that removes spaces as a post-processing step. The reasoning behind this is that it reduces the sequence length significantly with minimal information loss, and more closely mirrors existing models which have no explicit space information. Example tokenisations for the Unigram algorithms given the input ``This is an input sentence.'' are:

\begin{description}
    \small
    \item[Unigram] \_This, \_is, \_an, \_input, \_sentence, .
    \item[Unigram$'$] This, \_, is, \_, an, \_, input, \_, sentence, .
    \item[Unigram$'$ no spaces]  This, is, an, input, sentence, .
\end{description}

For each of the tokenisers, we pretrain RoBERTa (base) on the full text of English Wikipedia, and then finetune on downstream tasks, keeping all hyperparameters fixed, changing only the tokenisation algorithm used. For evaluation of the models in a general domain, we use the GLUE benchmark \cite{wang2018glue}, excluding WNLI. For evaluation in specifically handling complex words, we use the two Superbizarre topicality tasks \cite{hofmann2021superbizarre}, which require the binary classification of derivationally complex English words\footnote{We do not consider the Superbizarre sentiment task due to a higher proportion of uninformative words.}.

Over the whole of the English Wikipedia data, the sequence lengths for each of the tokenisation approaches are:

\begin{description*}
    \small
    \item[BPE] 3.72e+09
    \item[BPE$'$] 5.88e+09
    \item[BPE$'$ no spaces] 3.61e+09
    \item[Unigram] 3.68e+09
    \item[Unigram$'$] 5.94e+09
    \item[Unigram$'$ no spaces] 3.67e+09
\end{description*}

As in the evaluation in \autoref{table:morphology-comparison}, the modified models without spaces give shorter sequences than their default counterparts, with BPE$'$ without spaces giving the shortest mean sequence length. The difference in sequence lengths of the models means a difference in number of updates per epoch during pretraining. Hence, fixing the number of updates (and thus training time) will advantage models with shorter sequence lengths, especially disadvantaging the models that include spaces. Because of this, we perform two evaluations: one fixing the number of pretraining updates, and one fixing the number of pretraining epochs\footnote{In finetuning, the number of updates and epochs is equivalent for all models as one example is processed at a time. In pretraining, we follow the standard implementation of RoBERTa by taking contiguous sentences from the training data up to the max sequence length.}. 

Due to computational constraints, we only ran pretraining once for each model. For finetuning, we ran each experiment with 10 different seeds, reporting the mean development result and standard deviation. Results are shown in \autoref{table:results-fixed-updates} and \autoref{table:results-fixed-epochs} for fixed updates and fixed epochs, respectively. Full training procedure is given in \Cref{sec:app-training}.

\begin{table*}[ht!]
{
\renewcommand{\arraystretch}{1.2}
\begin{tabularx}{\linewidth}{|c*{7}{|>{\centering\arraybackslash}X}}
\hline
 & Epochs & GLUE & \multicolumn{2}{c|}{Superbizarre Reddit} & \multicolumn{2}{c|}{Superbizarre Arxiv} \\
  \cline{4-7}
& & & Dev & Test & Dev & Test \\ 
 \hline
DelBERT (supervised) & - & - & 69.6 & 70.1 & 73.1 & 72.3 \\ 
\hdashline
BPE & 27 & 81.6 & 66.8 & 66.6 & 71.1 & 70.2 \\
BPE$'$ & 16 & 79.2 & 66.6 & 66.2 & 70.3 & 69.3 \\
BPE$'$ no spaces & 28 & 81.7 & 67.2 & 66.9 & 70.9 & 70.0 \\
\hline
Unigram & 27 & 81.5 & 68.0 & 67.8 & 72.2 & 71.4 \\
Unigram$'$ & 16 &  78.4 & 68.2 & 68.2 & 72.5 & 71.6 \\
Unigram$'$ no spaces & 27 & 81.9 & \textbf{68.8} & \textbf{68.8} & \textbf{73.0} & \textbf{72.3} \\
\hline
\end{tabularx}
}
\caption{\label{table:results-fixed-updates} Finetuning results after pretraining for 100,000 updates. Shown are mean results across 10 seeds. Results that are significantly better than all others using a Welch's t-test ($ p < 0.05$) are shown in bold. More detailed results are given in \Cref{sec:app-detailed-results}. We include DelBERT \cite{hofmann2021superbizarre} as a supervised baseline, where the models are passed a morphological parse of the input.}
\end{table*}

\begin{table*}[ht!]
{
\renewcommand{\arraystretch}{1.2}
\begin{tabularx}{\linewidth}{|c*{7}{|>{\centering\arraybackslash}X}}
\hline
 & Updates & GLUE & \multicolumn{2}{c|}{Superbizarre Reddit} & \multicolumn{2}{c|}{Superbizarre Arxiv} \\
 \cline{4-7}
 &  &  & Dev & Test & Dev & Test \\ 
 \hline
DelBERT (supervised) & - & - & 69.6 & 70.1 & 73.1 & 72.3 \\ 
\hdashline
BPE & 109,761 & 81.5 & 67.1 & 66.8 & 71.0 & 70.1  \\
BPE$'$ & 177,845 & 79.5 & 66.8 & 66.5 & 70.5 & 69.8 \\
BPE$'$ no spaces & 106,485 & 81.5 & 67.1 & 67.1 & 70.8 & 70.1 \\
\hline
Unigram & 108,606 & 81.6 & 67.9 & 67.9 & 72.2 & 71.6 \\
Unigram$'$ & 179,909 & 79.1 & 68.3 & 68.3 & 72.5 & 71.8 \\
Unigram$'$ no spaces & 108,441 & 81.8 & \textbf{68.8} & \textbf{69.0} & \textbf{73.2} & \textbf{72.5} \\
\hline
\end{tabularx}
}

\caption{\label{table:results-fixed-epochs} Finetuning results after pretraining for 30 epochs. Shown are mean results across 10 seeds. Results that are significantly better than all others using a Welch's t-test ($ p < 0.05$) are shown in bold. More detailed results are given in \Cref{sec:app-detailed-results}. We include DelBERT \cite{hofmann2021superbizarre} as a supervised baseline, where the models are passed a morphological parse of the input.}
\end{table*}

On the Superbizarre datasets, we can see that Unigram outperforms BPE, with Unigram$'$ no spaces performing significantly better than all other models using a Welch's t-test ($ p < 0.05$), see \Cref{sec:app-significance-tests}. Note that DelBERT \cite{hofmann2021superbizarre}, a model which is passed the input segmented by a morphological algorithm, achieves 73.1 on the Arxiv dev set and 72.3 on the Arxiv test set, both worse than our (unsupervised) model, although DelBERT outperforms our best models on the Reddit task, achieving 69.6 and 70.1 on the dev and test sets, respectively.

On the mean GLUE benchmark, the modified models without spaces perform as well or better than their default counterparts, with Unigram$'$ performing the best when both updates and epochs are fixed. However, this result is not statistically significant (see \Cref{sec:app-significance-tests}), and over the individual GLUE tasks the best performing models vary, with high variances across seeds on some tasks due to the small dataset sizes (see \Cref{sec:app-detailed-results}). Since the GLUE tasks do not rely on handling complex words, a significant performance difference is probably not expected, but we see no drop in performance with the modified algorithms.

The modified models that include spaces perform poorly on the GLUE benchmark, even when the number of epochs is fixed rather than updates, meaning they are trained for $\sim$65\% more updates than the modified models without spaces. This suggests that this method of including spaces as additional tokens is suboptimal for general language tasks, though interestingly Unigram$'$ with spaces is the second best performing model across all Superbizarre datasets. The tokenisers themselves perform splitting on spaces as a first step, so additionally including spaces may be simply passing noise to the model for the masked language modelling task, especially due to the high frequency of spaces. This means the pretraining loss decreases rapidly due to space prediction, but plateaus earlier (see \Cref{sec:app-training}). Due to the much greater sequence lengths, the models that include spaces also discard examples that are too long during finetuning, which could lead to worse results. 

\section{Related Work}
There are previous works that have performed controlled extrinsic comparisons of existing subword tokenisation algorithms (BPE, Unigram, and WordPiece), and have provided results which we relate here to our own findings. 
\citet{galle2019investigating} investigates various compression algorithms for tokenisation, including BPE, and finds an inverse link between mean tokens per sentence and translation quality, hypothesising that the compression capability of BPE leads to its effectiveness in NLP tasks. In our experiments we find that Unigram$'$ outperforms BPE$'$ on the complex words tasks, and there to be no significant difference between them on the general language understanding (GLUE) tasks. This is despite Unigram$'$ having a longer sequence length, suggesting this factor is not wholly indicative of model performance. However, if we look at the results for fixed pretraining updates, we do see a slight negative correlation between sequence length and performance on the Superbizarre datasets, and a very strong negative correlation on the GLUE benchmark\footnote{Pearson correlations between -0.157 and -0.224 for the Superbizarre datasets, and -0.985 for GLUE.}, though this is skewed by the models including spaces performing very poorly. Intrinsically, we see a correlation (albeit weak) between sequence length and morphological correctness (\Cref{sec:intrinsic-evaluation}).
\citet{bostrom2020byte} compare Unigram and BPE, finding that Unigram generates more morphologically correct tokenisations and gives improved downstream task performance. Whilst we saw similar improvements in intrinsic performance, we were unable to replicate the performance difference on MNLI that they found, finding no significant difference in performance (see \Cref{sec:app-detailed-results}). We did not perform evaluation on the other two English datasets they used. 
\citet{hofmann-etal-2022-embarrassingly} corroborate these intrinsic results, additionally finding the morphological quality of WordPiece to lie in between that of BPE and Unigram, reflecting our own findings (\Cref{sec:intrinsic-evaluation}). 
\citet{wei2021training} perform comparison between byte-level BPE and byte-level Unigram, finding BPE to perform better than Unigram across seven languages on the XNLI dataset, which is contrary to our findings and those of \citet{bostrom2020byte} and \citet{hofmann-etal-2022-embarrassingly}.

There have also been some recent attempts to develop improved subword tokenisation methods. 
\citet{hofmann2021superbizarre} introduce DelBERT, which takes input words tokenised according to gold standard morphological references, with an unchanged vocabulary. They find that this improves performance on their Superbizarre datasets (\Cref{sec:extrinsic-evaluation}). \citet{hofmann-etal-2022-embarrassingly} also introduce FLOTA (Few Longest Token Approximation), which improves the performance of BERT, GPT-2, and XLNET at classifying ArXiv papers into their subareas from the title. \citet{yehezkel2022incorporating} introduce a context-aware tokeniser, SaGe, which they find improves performance over BPE on GLUE tasks, the Turkish subset of XLNI, and NER in both Turkish and English. 
There are also alternative subword tokenisation algorithms which have a history of use in machine translation tasks, including Morfessor \cite{creutz2002unsupervised} and its successors (\citealt{virpioja2013morfessor}, \citealt{gronroos2020morfessor}), and Dynamic Programming Encoding (DPE) \cite{he-etal-2020-dynamic}. (See \citealt{mielke2021between} for a more extensive review.)

For all of these approaches, spaces still occur as the first character of start-of-word tokens, and we believe this hinders performance: our alternative treatment of spaces could be combined with these algorithms, and the impact on performance investigated.

Finally, we note that \citet{wei2021training} experiment with different methods of handling spaces within their byte-level BPE algorithm which appear similar to those implemented here, although they find these alternatives perform worse than the default on XNLI. They do not release code for their experiments so unfortunately we are unable to make a controlled comparison. 

\section{Conclusion and Future Work}
We hypothesise that problems with current tokenisation algorithms arise from allowing tokens to include spaces, and thus experiment with an alternative tokenisation approach where spaces are always treated as individual tokens. We find that this leads to improved performance on NLP tasks involving complex words, whilst having no detrimental effect on performance in general natural language understanding tasks. Whilst our work focuses on BPE and Unigram, our modifications can be applied to any existing subword tokenisation algorithm, including WordPiece, and hence to any transformer-based model. Also, although our experiments have only been in English, the algorithms used are unsupervised and language-independent and our results should extend to other languages.

Our best-performing models use lossy tokenisation (removing the space tokens as a post-processing step), which may not be ideal for all tasks. We did not perform evaluation on sequence-to-sequence tasks, and indeed the subword tokenisation algorithms discussed here were introduced in the field of NMT, where space information needs to be generated in the output. Future work could thus look at alternative methods for including space information that maintain the performance gains seen here whilst keeping tokenisation lossless.

\subsection*{Acknowledgements}
This work was partially supported by the CDT in Speech and Language Technologies and their Applications funded by UKRI (grant number EP/S023062/1) and the UK EPSRC grant EP/T02450X/1. We thank the reviewers for their helpful feedback.

\section*{Limitations}
The finetuning tasks investigated in this paper are all sequence classification, which is a significant limitation of the evaluation. In order to definitively compare our modified tokenisation algorithms with the defaults, a more thorough evaluation across many types of encoder-architecture NLP tasks would be required (e.g. token classification, question answering, multiple-choice). It is also worth noting that the Superbizarre dataset consists of entries constructed using elements from BERT’s WordPiece vocabulary. For their purposes, this is a benefit as it does not unfairly disadvantage BERT, but for our purposes it limits the generality of the results obtained. 
In this paper, we have chosen a single vocabulary size for all of our evaluation, which limits the robustness of our results. For the intrinsic evaluation, a range of vocabulary sizes could be chosen and evaluated. For extrinsic evaluation, we are limited by the computational expense of pretraining language models, but it is important to note that we don't know how our results will change if the vocabulary size is altered. It would also be beneficial to look at how our modified tokenisers work on morphologically rich languages, and in a multilingual setting, which would further increase the robustness of the results.

\bibliographystyle{acl_natbib}
\bibliography{custom}

\clearpage

\appendix

\section{Training Details}
\label{sec:app-training}
Hyperparameters for tokenisation, pretraining, finetuning are shown in \autoref{table:hyperparameters-tokeniser}, \autoref{table:hyperparameters-pretraining} and \autoref{table:hyperparameters-finetuning}, respectively.
We did not use stochastic tokenisation (BPE-dropout or subword regularisation).

\begin{table}[ht]
\small
\centering
\resizebox{\linewidth}{!}{%
\begin{tabular}{|p{0.35\linewidth}|p{0.65\linewidth}|}
\hline
Implementation & SentencePiece \cite{kudo2018sentencepiece} \\
Vocabulary Size & 16000 \\
BPE-dropout & 0 \\
Unigram Subword Regularisation & 0 \\ 
\hline
\end{tabular}
}
\caption{\label{table:hyperparameters-tokeniser} Hyperparameters for tokenisation.}
\end{table}

\begin{table}[ht]
\small
\centering
\resizebox{\linewidth}{!}{%
\begin{tabular}{|p{0.35\linewidth}|p{0.65\linewidth}|}
\hline
Implementation & fairseq \cite{ott2019fairseq} \\
Architecture & RoBERTa (base) \cite{liu2019roberta} \\
Precision & 16 bit\\
Optimizer &ADAM \cite{kingma2014adam}, $\epsilon=$1e-6, $\beta=(0.9,0.98)$\\ 
Sequence length & 512\\
Learning rate scheduler & Linear warm-up for 10000 updates to 5e-4, then reduce to 1e-4 upon increased training loss at epoch \\ 
Training for & 100000 updates / 30 epochs  \\
Batch size & 2048  \\ 
Dropout & 0.1 \\ 
Attention Dropout & 0.1 \\ 
Weight Decay & 0.01 \\ 

\hline
\end{tabular}
}
\caption{\label{table:hyperparameters-pretraining} Hyperparameters for pretraining.}
\end{table}

\begin{table}[ht]
\small
\centering
\resizebox{\linewidth}{!}{%
\begin{tabular}{|p{0.35\linewidth}|p{0.65\linewidth}|}
\hline
Implementation & fairseq \cite{ott2019fairseq} \\
Architecture & RoBERTa (base) \cite{liu2019roberta} \\
Precision & 16 bit \\
Optimizer & ADAM \cite{kingma2014adam}, $\epsilon=$1e-6, $\beta=(0.9,0.98)$ \\ 
Sequence length & 512 \\
Learning rate scheduler & Linear warm-up to 2e-3 for 6\% of updates, then linear decay to 0 \\ 
Training for & 20 epochs \\
Batch size & 32 \\ 
Dropout & 0.1 \\ 
Attention Dropout & 0.1 \\ 
Weight Decay & 0.01 \\ 

\hline
\end{tabular}
}
\caption{\label{table:hyperparameters-finetuning} Hyperparameters for finetuning.}
\end{table}

\subsection{Pretraining}
Pretraining was run on 8 NVIDIA Tesla V100s.
We ran pretraining on the text of English Wikipedia.
A Wikipedia dump was processed with the Python package WikiExtractor\footnote{\url{https://github.com/attardi/wikiextractor/}}, and then split into sentences using BlingFire\footnote{\url{https://github.com/microsoft/BlingFire}}. 
In order to perform a fair comparison across models, we removed all sentences with sequence lengths longer than 510 when tokenised with the modified models including spaces. However, this was a very small amount of the data ($\sim$0.002\%) and would therefore have a negligible effect on performance. 

\begin{figure*}
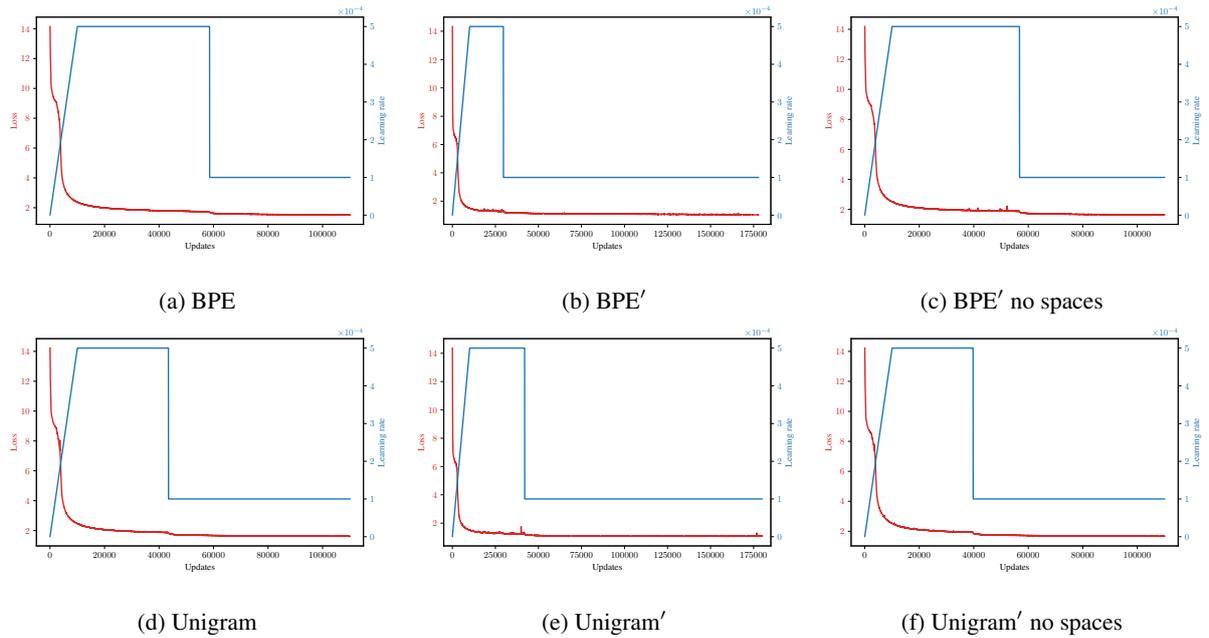


    \begin{subfigure}[t]{0.33\textwidth}
    \begin{center}
        \resizebox{.99\linewidth}{!}{\input{bpe_spm3.pgf}}
    \end{center}
    \caption{BPE}
    \end{subfigure}
    \begin{subfigure}[t]{0.33\textwidth}
    \begin{center}
        \resizebox{.99\linewidth}{!}{\input{bpe_spm1.pgf}}
    \end{center}
    \caption{BPE$'$}
    \end{subfigure}
    \begin{subfigure}[t]{0.33\textwidth}
    \begin{center}
        \resizebox{.99\linewidth}{!}{\input{bpe_spm2.pgf}}
    \end{center}
    \caption{BPE$'$ no spaces}
    \end{subfigure}
    \begin{subfigure}[t]{0.33\textwidth}
    \begin{center}
        \resizebox{.99\linewidth}{!}{\input{unigram_spm3.pgf}}
    \end{center}
    \caption{Unigram}
    \end{subfigure}
    \begin{subfigure}[t]{0.33\textwidth}
    \begin{center}
        \resizebox{.99\linewidth}{!}{\input{unigram_spm1.pgf}}
    \end{center}
    \caption{Unigram$'$}
    \end{subfigure}
    \begin{subfigure}[t]{0.33\textwidth}
    \begin{center}
        \resizebox{.99\linewidth}{!}{\input{unigram_spm2.pgf}}
    \end{center}
    \caption{Unigram$'$ no spaces}
    \end{subfigure}
    
    \caption{\label{table:loss-curves} Pretraining loss curves for the six models.}
\end{figure*}

Loss curves are shown in \autoref{table:loss-curves}.

\subsection{Finetuning}
Finetuning was run on a single NVIDIA Tesla V100.
All finetuning experiments were ran with a batch size of 32, and a peak learning rate of 2e-3 with linear warm-up for 6\% of updates, then linear decay to 0. All other parameters were kept the same as for pretraining. Experiments were ran for 20 epochs, and the best performing epoch was taken, with 10 random seeds per model. For the Superbizarre datasets, we took the best performing epoch for each seed on the dev set and evaluated it on the test set.

\section{Detailed Results}
\label{sec:app-detailed-results}
Detailed results are shown in \autoref{table:detailed-results-fixed-updates} and \autoref{table:detailed-results-fixed-epochs} for fixed pretraining updates and fixed pretraining epochs, respectively. The standard deviations on the mean GLUE score are calculated assuming zero covariance between tasks.

\begin{table*}[ht!]
\tiny
{
\renewcommand{\arraystretch}{1.5}
\begin{tabularx}{\linewidth}{c*{16}{>{\centering\arraybackslash}X}}
\hline
 & Epochs & \multicolumn{10}{c|}{GLUE} & \multicolumn{2}{c|}{Superbizarre Reddit} & \multicolumn{2}{c|}{Superbizarre Arxiv} \\
& & MRPC & CoLA & STS-B & RTE & SST-2 & QQP & QNLI & MNLI-m & MNLI-mm & Mean & Dev & Test & Dev & Test \\ 
 \hline
BPE & 27 & 84.5 (0.8) & 55.4 (2.5) & 87.1 (0.3) & 68.6 (2.7) & 91.6 (0.4) & 89.7 (0.1) & 91.3 (0.2) & 83.1 (0.2) & 83.5 (0.3) & 81.6 (1.3) & 66.8 (0.8) & 66.6 (0.9) & 71.1 (0.2) & 70.2 (0.2) \\
BPE$'$ & 16 & 83.0 (1.0) & 48.9 (2.9) & 86.0 (0.2) & 59.5 (1.9) & 91.6 (0.4) & 89.2 (0.1) & 90.7 (0.3) & 81.6 (0.2) & 82.3 (0.1) & 79.2 (1.2) & 66.6 (0.2) & 66.2 (0.2) & 70.3 (0.1) & 69.3 (0.2) \\
BPE$'$ no spaces & 28 & 84.4 (0.6) & 54.4 (1.4) & 87.0 (0.2) & 70.3 (0.8) & 92.2 (0.5) & 89.7 (0.1) & 91.1 (0.2) & 83.1 (0.2) & 83.2 (0.2) & 81.7 (0.6) & 67.2 (0.2) & 66.9 (0.2) & 70.9 (0.1) & 70.0 (0.2) \\
\hline
Unigram & 27 & 85.0 (1.2) & 52.3 (1.4) & 87.3 (0.2) & 69.8 (1.9) & 91.7 (0.5) & 89.5 (0.1) & 91.9 (0.4) & 83.1 (0.2) & 83.1 (0.2) & 81.5 (0.9) & 68.0 (0.2) & 67.8 (0.3) & 72.2 (0.3) & 71.4 (0.2) \\
Unigram$'$ & 16 & 83.3 (0.6) & 39.5 (15.4) & 84.8 (0.4) & 64.0 (1.8) & 91.3 (0.4) & 89.1 (0.1) & 89.8 (0.3) & 81.4 (0.2) & 82.1 (0.2) & 78.4 (5.2) & 68.2 (0.4) & 68.2 (0.3) & 72.5 (0.2) & 71.6 (0.3) \\
Unigram$'$ no spaces & 27 & 85.2 (1.4) & 54.6 (1.4) & 87.8 (0.3) & 71.1 (1.5) & 91.6 (0.4) & 89.5 (0.1) & 91.3 (0.3) & 83.0 (0.2) & 83.1 (0.2) & 81.9 (0.9) & 68.8 (0.1) & 68.8 (0.3) & 73.0 (0.2) & 72.3 (0.3) \\
\hline
\end{tabularx}
}
\caption{\label{table:detailed-results-fixed-updates} Full finetuning results after pretraining for 100000 updates. Shown are mean dev set results across 10 seeds, with standard deviations in parentheses.}
\end{table*}

\begin{table*}[ht!]
\tiny
{
\renewcommand{\arraystretch}{1.5}
\begin{tabularx}{\linewidth}{c*{16}{>{\centering\arraybackslash}X}}
\hline
 & Updates & \multicolumn{10}{c|}{GLUE} & \multicolumn{2}{c|}{Superbizarre Reddit} & \multicolumn{2}{c|}{Superbizarre Arxiv}  \\
 &  & MRPC & CoLA & STS-B & RTE & SST-2 & QQP & QNLI & MNLI-m & MNLI-mm & Mean & Dev & Test & Dev & Test \\ 
 \hline
BPE & 109761 & 84.4 (0.8) & 53.5 (1.7) & 87.2 (0.2) & 68.7 (0.9) & 91.8 (0.3) & 89.7 (0.1) & 91.4 (0.2) & 83.1 (0.2) & 83.5 (0.3) & 81.5 (0.7) & 67.1 (0.2) & 66.8 (0.3) & 71.0 (0.2) & 70.1 (0.3) \\
BPE$'$ & 177845 & 83.2 (1.1) & 48.9 (1.4) & 86.6 (0.2) & 60.0 (2.6) & 92.0 (0.2) & 89.2 (0.0) & 90.7 (0.3) & 82.2 (0.2) & 82.9 (0.2) & 79.5 (1.1) & 66.8 (0.3) & 66.5 (0.1) & 70.5 (0.1) & 69.8 (0.2)\\
BPE$'$ no spaces & 106485 & 85.0 (0.6) & 53.4 (0.9) & 86.9 (0.3) & 69.1 (0.6) & 92.0 (0.4) & 89.5 (0.1) & 91.2 (0.3) & 83.2 (0.2) & 83.2 (0.2) & 81.5 (0.5) & 67.1 (0.2) & 67.1 (0.3) & 70.8 (0.2) & 70.1 (0.2)  \\
\hline
Unigram & 108606 & 84.8 (0.9) & 53.1 (2.3) & 87.4 (0.2) & 70.1 (1.8) & 91.6 (0.3) & 89.6 (0.1) & 91.3 (0.5) & 83.0 (0.1) & 83.2 (0.2) & 81.6 (1.0) & 67.9 (0.2) & 67.9 (0.3) & 72.2 (0.1) & 71.6 (0.1) \\
Unigram$'$ & 179909 & 82.0 (0.9) & 45.9 (2.0) & 84.7 (0.2) & 64.9 (1.5) & 91.5 (0.3) & 89.0 (0.1) & 90.1 (0.2) & 81.5 (0.1) & 82.0 (0.1) & 79.1 (0.9) & 68.3 (0.5) & 68.3 (0.4) & 72.5 (0.2) & 71.8 (0.3) \\
Unigram$'$ no spaces & 108441 & 84.8 (0.8) & 54.5 (1.9) & 87.8 (0.2) & 70.0 (1.8) & 91.5 (0.3) & 89.6 (0.1) & 91.5 (0.2) & 83.2 (0.1) & 83.2 (0.2) & 81.8 (0.9) & 68.8 (0.2) & 69.0 (0.2) & 73.2 (0.2) & 72.5 (0.2) \\
\hline
\end{tabularx}
}

\caption{\label{table:detailed-results-fixed-epochs} Full finetuning results after pretraining for 30 epochs. Shown are mean dev set results across 10 seeds, with standard deviations in parentheses.}
\end{table*}

\section{Significance Tests}
\label{sec:app-significance-tests}
Here we give full Welch's t-test results comparing the best performing model to all the others for each dataset, shown in \autoref{table:welchs-tests-fixed-updates} and \autoref{table:welchs-tests-fixed-epochs} for fixed pretraining updates and fixed pretraining epochs, respectively. 

\begin{table*}[ht!]
{
\renewcommand{\arraystretch}{1.2}
\begin{tabularx}{\linewidth}{c*{6}{>{\centering\arraybackslash}X}}
\hline
  & GLUE & \multicolumn{2}{c|}{Superbizarre Reddit} & \multicolumn{2}{c|}{Superbizarre Arxiv} \\
 &    & Dev & Test & Dev & Test \\ 
 \hline
BPE  & 0.61 & 2.15e-05 & 1.34e-05 & 5.70e-15 & 5.26e-13 \\
BPE$'$ & 2.7e-05 & 1.50e-16 & 5.26e-14 & 3.82e-17 & 8.61e-15\\
BPE$'$ no spaces & 0.58 & 7.22e-14 & 9.04e-12 & 1.75e-15 & 5.54e-14 \\
\hline
Unigram & 0.36 & 2.27e-08 & 1.69e-06 & 5.15e-07 & 1.11e-07\\
Unigram$'$ & 6.0e-02 & 6.22e-04 & 7.83e-05 & 1.74e-05 & 6.05e-06  \\
\hline
\end{tabularx}
}
\caption{\label{table:welchs-tests-fixed-updates} P values for Welch's t-test comparing Unigram$'$ no spaces to other models for fixed pretraining updates. }
\end{table*}

\begin{table*}[ht!]
{
\renewcommand{\arraystretch}{1.2}
\begin{tabularx}{\linewidth}{c*{6}{>{\centering\arraybackslash}X}}
\hline
  & GLUE & \multicolumn{2}{c|}{Superbizarre Reddit} & \multicolumn{2}{c|}{Superbizarre Arxiv} \\
 &    & Dev & Test & Dev & Test \\ 
 \hline
BPE  & 0.41 & 1.47e-13 & 2.25e-13 & 2.77e-15 & 1.48e-13 \\
BPE$'$ & 8.72e-05 & 1.19e-12 & 7.84e-16 & 3.46e-16 & 4.53e-14 \\
BPE$'$ no spaces & 0.41 & 1.28e-12 & 2.01e-15 & 1.21e-15 & 2.35E-12\\
\hline
Unigram & 0.66 & 2.92e-08 & 1.19e-09 & 3.96e-09 & 8.78e-08  \\ 
Unigram$'$ &  2.8e-06 & 1.45e-02 & 1.69e-06 & 1.55e-06 & 3.90e-04 \\

\hline
\end{tabularx}
}
\caption{\label{table:welchs-tests-fixed-epochs} P values for Welch's t-test comparing Unigram$'$ no spaces to other models for fixed pretraining epochs. }
\end{table*}

\end{document}